\documentclass[conference]{IEEEtran}
\IEEEoverridecommandlockouts
\usepackage{cite}
\usepackage{amsmath,amssymb,amsfonts}
\usepackage{algorithmic}
\usepackage{graphicx}
\usepackage{textcomp}
\usepackage{xcolor}
\def\BibTeX{{\rm B\kern-.05em{\sc i\kern-.025em b}\kern-.08em
    T\kern-.1667em\lower.7ex\hbox{E}\kern-.125emX}}
\usepackage[caption=false]{subfig}
\usepackage{makecell}
\usepackage{amsmath}
\usepackage{bm}
\usepackage{graphicx}
\usepackage{hyperref}
\usepackage{xspace}
\usepackage{amsmath}
\usepackage{amsfonts}
\usepackage{algorithm}
\usepackage{paralist}
\usepackage[bottom]{footmisc}
\usepackage{mathtools}
\usepackage{booktabs}
\usepackage{enumitem}

\newcommand{\fgr}[3][\relax]{%
	\begin{figure}[htp]%
		\centering
		\includegraphics[#2]{#3}%
		\vspace{-0.1in}
		\ifx\relax#1\else\caption{{#1}}\fi
	\end{figure}%
}

\newcommand{\hide}[1]{}

\newcommand{\method}{{\sc ELECT}\xspace}

\newcommand{\metaod}{{\sc MetaOD}\xspace}

\newcommand{\cbit}{\begin{compactitem}}
	\newcommand{\ceit}{\end{compactitem}}
\newcommand{\cben}{\begin{compactenum}}
	\newcommand{\ceen}{\end{compactenum}}

\newcommand{\bal}{\begin{align}}
\newcommand{\ean}{\end{align}}
\newcommand{\bit}{\begin{itemize}}
\newcommand{\eit}{\end{itemize}}
\newcommand{\ben}{\begin{enumerate}}
\newcommand{\een}{\end{enumerate}}
\newcommand{\beq}{\begin{equation}}
\newcommand{\eeq}{\end{equation}}

\newcommand{\R}{\mathbb{R}}

\DeclareMathAlphabet{\mathbcal}{OMS}{cmsy}{b}{n}

\newcommand{\mDt}{\mathbcal{D}_{\text{train}}}
\newcommand{\mDs}{\mathcal{D}_{\text{test}}}
\newcommand{\mD}{\mathcal{D}}

\newcommand{\mM}{\mathcal{M}}
\newcommand{\mN}{\mathcal{N}}

\newcommand{\bXs}{\mathbf{X}_{\text{test}}}
\newcommand{\bX}{\mathbf{X}}
\newcommand{\by}{\mathbf{y}}

\newcommand{\bmij}{\mathbf{m}_{i,j}}
\newcommand{\bmijp}{\mathbf{m}_{i,j'}}

\newcommand{\mO}{\mathcal{O}}

\newcommand{\bP}{\mathbf{P}}


\DeclareMathOperator*{\argmax}{arg\,max}

\newtheorem{problem}{Problem}

\newtheorem{criterion}{Criterion}

\newtheorem{lemma}{Lemma}

\makeatother

\newcommand\crule[3][black]{\textcolor{#1}{\rule{#2}{#3}}}

\definecolor{aliceblue}{rgb}{0.73,0.83,0.91}
\definecolor{alicegreen}{rgb}{0.74,0.88,0.74}
\definecolor{alicered}{rgb}{0.945,0.738,0.738}

\begin{document}


\title{Toward  Unsupervised Outlier Model Selection
}

\vspace{-0.1in}
\author{
\IEEEauthorblockN{
Yue Zhao}
\IEEEauthorblockA{
\textit{Carnegie Mellon University}\\
zhaoy@cmu.edu}
\and
\IEEEauthorblockN{
Sean Zhang}
\IEEEauthorblockA{
\textit{Carnegie Mellon University}\\
xiaoronz@alumni.cmu.edu}
\and
\IEEEauthorblockN{
Leman Akoglu}
\IEEEauthorblockA{
\textit{Carnegie Mellon University}\\
lakoglu@andrew.cmu.edu}

}
\vspace{-0.3in}

\maketitle

\begin{abstract}
Today there exists no shortage of outlier detection algorithms in the literature, yet the complementary and  critical problem of unsupervised outlier model selection (UOMS) is vastly understudied.
In this work, we propose \method, a new approach to select an effective candidate model, i.e. an outlier detection algorithm and its hyperparameter(s), to employ on a {\em new dataset without any labels}.
At its core, \method is based on meta-learning; transferring prior knowledge (e.g. model performance) on historical datasets that are \textit{similar} to the new one
to facilitate UOMS. Uniquely, it employs a dataset similarity measure that is performance-based, 
which is more direct and goal-driven than other measures used in the past.
\method adaptively searches for similar historical datasets, as such, it can serve an output on-demand, being able to accommodate varying time budgets. Extensive experiments show that \method significantly outperforms a wide range of basic UOMS baselines, including no model selection (always using the same popular model such as iForest) as well as more recent selection strategies based on  meta-features. 
\vspace{-0.1in}

\end{abstract}


\vspace{-0.075in}
\section{Introduction}
\label{sec:intro}

Outlier detection (OD) aims to identify data points deviating from the main data generating distribution, with numerous applications such as
intrusion detection, 
anti-money laundering,  
rare disease detection,  
to name a few. Over the years, a large body of new detection algorithms are proposed in the literature \cite{aggarwal2015outlier},
including the most recent surge of deep learning \cite{pang2021deep}.

While there is no shortage of OD algorithms today, a key question remains untackled: \textit{which algorithm and values of hyperparameter(s)} (HP) (together referred to as OD model or outlier model) to use on a given new dataset (or detection task)? 
Broadly known as the \textit{model selection} problem, this question is at the heart of (machine) learning with direct implications on performance and generalization \cite{raschka2018model}, and OD is no exception \cite{Campos2016,Bahri2022automl}. 
As the ``no free lunch'' theorem \cite{DolpertM97} implies, there exists no ``winner'' outlier model that 
excels across all tasks \cite{han2022adbench}, especially provided that OD is employed in a wide variety of domains. In fact, the best OD model can vary even on tasks within the same domain; for example, distinct OD algorithms are reported to outperform in three separate articles on network intrusion detection \cite{DBLP:journals/corr/MarteauSB17,DBLP:conf/sdm/LazarevicEKOS03,DBLP:conf/socpar/KumarK16}.
Moreover, the sensitivity of OD algorithms to their HP settings is recognized in several evaluation studies \cite{journals/sigkdd/AggarwalS15,goldstein2016comparative,Campos2016}.
For instance, the literature reports that by varying the number of nearest neighbors in local outlier factor (LOF) \cite{breunig2000lof} while keeping the other conditions the same, up to 10$\times$ performance
difference is observed in some datasets \cite{DBLP:journals/corr/abs-2009-10606}. 
All of these works show the fact that OD model selection is critical and inevitable.

Moving beyond designing yet-another detection algorithm, 
it is exactly our goal in this work to systematically address the unsupervised outlier model selection (UOMS) problem, which involves selecting an OD algorithm as well as its hyperparameter(s) for a given new OD dataset.
While essential, the problem is notoriously hard for: (i) model evaluation (say, on hold-out data) is infeasible due to the lack of labels, and (ii) model comparison is inapplicable as there is no universal loss/objective criterion applicable to all OD algorithms. 
Notably, even if they were available, using labels in OD model selection could be challenged as the available labels may be too limited to be  comprehensive/high-coverage, and in turn the selected model based on known labels may not be suited 
to unknown/emerging types of outliers in deployed systems.

{\bf Prior Work.~}
In supervised learning, model selection can be done via performance evaluation of each trained model on a labeled hold-out set, either by simply searching among models defined over a static grid or via more sophisticated dynamic search techniques such as bandit-based strategies \cite{li2017hyperband} and Bayesian hyperparameter optimization \cite{hutter2011sequential,thornton2013auto} -- the latter of which are effectively used in the AutoML literature \cite{he2021automl}. We remark that these techniques cannot (at least directly) be used in UOMS
as they require model evaluation using ground-truth labels. 
There exist some recent work on automating OD by Li {\em et al.} \cite{li2020autood,li2020pyodds}, which however also relies on labels, inheriting the aforementioned limitations.
Most recently, a meta-learning
based \textit{unsupervised} approach is proposed by Zhao {\em et al.} \cite{DBLP:journals/corr/abs-2009-10606} where similarity among the input task and historical OD tasks is leveraged, which is measured based on meta-features, i.e., summary statistics of a dataset. 
{Our present work is inspired by Zhao {\em et al.}'s work, yet takes a distinct
approach for quantifying task similarity.}
See discussion of related work in \S\ref{sec:related_work}.

{\bf Present Work.~}
{The primary motivation behind this work is to use \textit{performance-driven similarity} for  quantifying the resemblance 
among the input task and historical OD tasks in meta-learning.
Building upon this, we propose \method, an iterative meta-learning approach for unsupervised outlier model selection. In principle, meta-learning carries over prior ``experience'' from a database of historical tasks to facilitate the learning on a new task, provided that the new task resembles at least some of the historical ones. 
As we aim to carry over the information of prior performance of OD models on historical tasks to 
effectively select a high-performance model for an input task, we argue that the most suitable measure of task similarity is \textit{the similarity of model performances on two tasks}.}
Of course, model performances are unknown and cannot be directly evaluated on the new, unsupervised task. This is where we employ meta-learning by training a supervised predictor on the historical tasks that maps internal information of trained models 
(without using any labels) to their actual performance. 
We carefully and adaptively search for similar historical tasks and select a model that achieves high performance on those, without training too many models on the input task, such that UOMS incurs negligible computational overhead.
We find it important to remark that UOMS precedes OD (say, by the selected model), and that  
 \method is strictly a model selection technique other than a new OD algorithm. Our contributions:

\begin{itemize}
[leftmargin=*]

\setlength\itemsep{0.025in}
\item {\bf New UOMS Method.}
We introduce \method, a novel approach to unsupervised OD model selection based on meta-learning. It capitalizes on historical OD tasks with labels to select an effective, high-performance model for a new task without any labels. 

\item {\bf Performance-driven Task Similarity.} The key mechanism behind meta-learning is to effectively identify and transfer knowledge from \textit{similar} historical tasks to the new task. Unlike prior work \cite{DBLP:journals/corr/abs-2009-10606} that relies on handcrafted meta-features to quantify task similarity, \method takes a direct and goal-driven approach, and  deems two tasks similar if OD models \textit{perform} similarly on both tasks.

\item {\bf Unsupervised Adaptive Model Search.}
As ground-truth labels are unavailable for a new task, \method learns (during meta-training) to quantify performance based on \textit{internal} model performance measures. Moreover, it carefully decides which model to train on the new task iteratively, keeping as small as possible the total number of models trained before selection, thereby reducing computation time. Moreover, it is an \textit{anytime} algorithm that can output users a selected model at any time they choose to stop it.
\item {\bf Effectiveness.} Through extensive experiments on two testbeds, we show that \textit{selecting} a model by \method is significantly better than employing popular models like iForest
as well as all meta-feature baselines, including the SOTA MetaOD ($p=0.0016$), in the controlled testbed. 
\end{itemize}

\noindent {\bf Reproducibility.} To foster future work on UOMS, we fully open-source \method at \url{https://github.com/yzhao062/ELECT}.

\section{Problem Statement \& Overview}
\label{sec:prelim}

\subsection{Problem Statement}
We consider the model selection problem for unsupervised outlier detection (OD), which we refer to as UOMS (unsupervised outlier model selection) hereafter. 
Given a new dataset, \textit{without any labels}, the problem is to 
select a model -- jointly specified by both ($i$) a detector/OD algorithm and ($ii$) the values for its associated hyperparameter(s) (HP). The former is a discrete choice, given the finite set of existing detection algorithms.
The latter is continuous, and hence induces infinitely many candidate models. 

Continuous-space HP optimization is often addressed with iterative search strategies, such as particle swarm optimization and Bayesian optimization (BO) (see \cite{feurer2019hyperparameter,yang2020hyperparameter}), which adaptively and efficiently navigate the HP configuration space.
For example, BO decides which configuration to evaluate next
based on  previously evaluated configurations from prior rounds. An explore-exploit criterion guides this decision, trading off between (exploit) searching nearby high-performance configurations for local improvement  and (explore) wandering off to unexplored/uncertain regions of the space to improve the estimation of the global performance landscape.
These types of approaches are challenging to undertake for the UOMS problem, where model evaluation cannot be performed reliably due to the lack of ground-truth labels.

Therefore, in this work, we simplify and make the search space more tractable by pre-specifying the set of OD models to choose from.
Specifically we define a finite pool of models, denoted $\mM = \{M_1, \ldots, M_m\}$, by discretizing the HP space for each candidate detector.
Each model $M \in \mM$ here is a $\{$\texttt{detector}, \texttt{configuration}$\}$ pair, where the \texttt{configuration} depicts a specific setting of the HP values for the \texttt{detector}; e.g. $\{$\texttt{LOF}, \texttt{k=10}$\}$. 
Then, the UOMS problem is stated as follows:

\begin{problem}[{Unsupervised Outlier Model Section (UOMS)}]
\textit{\em\bf Given} a new unsupervised OD task, i.e. an input dataset 
$\mD_{\text{test}} = (\bXs, \emptyset)$ without any labels, and a finite candidate model set $\mM$; \textit{\em\bf Select} an effective model $M \in \mM$ to employ.
\end{problem}

\subsection{Proposed \method: Overview}
\label{subsec:overview}

At the heart of our proposed approach to UOMS lies \textit{meta-learning}, where the underlying principle is to transfer useful information from historical tasks to a new task. 
To this end, \method takes as input
a set of historical OD datasets 
$\mDt = \{\mD_1,\ldots,\mD_n\}$, namely, a meta-train database with ground-truth labels where $\{\mD_i = (\bX_i,\by_i)\}_{i=1}^n$
to compute:
\begin{itemize}[leftmargin=*]
\item the historical output scores of each candidate model $M_j \in \mM$ on each meta-train dataset $\mD_i \in \mDt$, where $\mO_{i,j} := M_j(\mD_i)$ refers to the $j$-th model's output outlier scores for the points in the $i$-th meta-train dataset $\mD_i$; 
and 
\item the historical performances matrix $\bP\in \R^{n\times m}$  of $\mM$ on $\mDt$, where
$\bP_{i,j}$ depicts model $M_j$'s performance\footnote{Area under the precision-recall curve (AUCPR, a.k.a. Average Precision or AP); can be substituted with any other accuracy measure of interest.}
on meta-train dataset $\mD_i$. 
\end{itemize}

\method consists of two phases.
In a nutshell, during the \textbf{(meta-)training phase (offline)}, it  learns information necessary to quantify similarity between two tasks based on $\mDt$.
It uses this information during the \textbf{(outlier) model selection phase (online)}  to
identify similar meta-train tasks to a new input task $\mDs$ and
chooses a model without using any labels.

\begin{figure}[!t]
\centering
	\includegraphics[width=\linewidth]{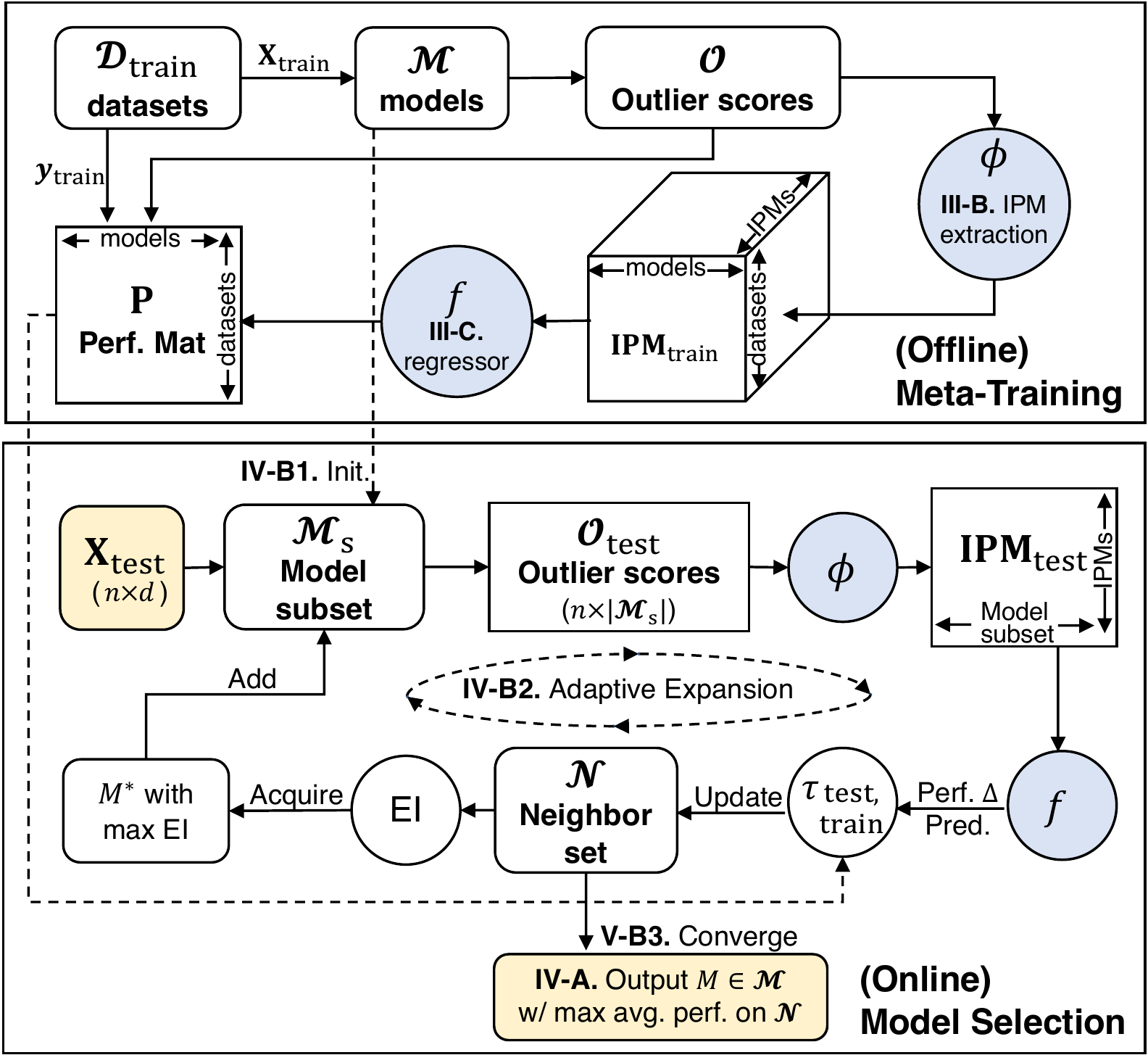}
	\vspace{-0.2in}
	\caption{
	Main 
	steps during the Offline and Online phases of \method.
	\label{fig:flowchart}
	}
	\vspace{-0.25in}
\end{figure}

Next, we present a high-level description of these phases.
Fig. \ref{fig:flowchart} illustrates the key steps in each phase.

\subsubsection{\textbf{(Offline) Meta-training Overview}}
Given historical tasks $\mDt$ and the new task $\mDs$ for model selection,
the main idea is to first identify a subset of tasks $\mN \subset \mDt$ that are \textit{similar} to $\mDs$, and then choose the model that performs the best on average on those ``neighbor'' tasks. Thus,  a key ingredient of \method is an effective {task similarity} measure.
Distinctly, it utilizes \textbf{performance-driven task similarity}: 
two tasks are 
similar if the performance rank-ordering of all the models 
on each task is similar. 

Of course, the similarity of $\mDs$ to
meta-train datasets cannot be computed based on the ground-truth model performances on $\mDs$ given that labels are unavailable for the input task (obviously, that would void the model selection problem altogether).
A core idea of \method is to learn to estimate true performance from \textbf{internal performance measures} (IPMs) during meta-training. IPMs are unsupervised signals that are solely based on the input points and/or a given model's output (e.g. outlier scores) 
that can be used to compare two models \cite{goix2016evaluate,marques2020internal}. However, they are noisy/weak indicators of performance \cite{DBLP:journals/corr/abs-2104-01422}. \method makes the best use of these weak signals by \textit{learning} to regress the IPMs of two models onto their true performance difference.

\subsubsection{\textbf{(Online) Model Selection Overview}}
In the (online) model selection phase, given $\mDs$, we can build/run each candidate model on $\bXs$, extract the corresponding IPMs,
 and use the regressor (from meta-training) to estimate pairwise model performance differences for quantifying $\mDs$'s similarity to each meta-train dataset based on the all-pairs rank similarity of the models.
 However, building and internally evaluating {\em all} of the candidate models would be computationally expensive. 
 
 To avoid training all the models at test time, a core idea behind \method is to quantify similarity between $\mDs$ and meta-train tasks based on only
a small \textit{subset} of the models, 
 which is carefully and adaptively picked and expanded. 
 At each iteration, this strategy picks the next model to train on $\mDs$ with the best explore-exploit trade-off, where exploration (picking a model with high uncertainty in performance estimate) helps refine similarity and thereby update the set of neighbor tasks, and exploitation (picking a model with high-performance estimate) ensures that the neighbor tasks share similarity w.r.t. the well-performing models.
 Ultimately, the goal is to effectively identify similar historical tasks based on as few models run on $\mDs$ as possible. Upon convergence, the model with the largest average performance on the neighbor tasks (from the latest iteration) is reported as the selected model for $\mDs$. 
 
We present the technical details of \method's (offline) meta-training and (online) model selection phases in \S \ref{sec:method}  and \S \ref{sec:online_model_selection}.
Additional implementation details are given in Appx. \ref{subsec:train_details} \& \ref{subsec:prediction_details}.

\section{\method: Meta-training (Offline)}
\label{sec:method}
The main component of meta-training is  
to learn, using historical (meta-train) tasks, how to quantify performance-based task similarity from imperfect indicators of performance, namely internal performance measures (IPMs).
In the following, we introduce our task similarity measure (\S\ref{subsec:sim}), specific IPMs used by \method (\S\ref{ssec:ipm}), and how to do the mapping between the two (\S\ref{ssec:performance_predictor}).  

\subsection{Performance-driven Task Similarity}
\label{subsec:sim}

Meta-learning carries the  prior experience on  historical tasks over to a new task, provided that the latter resembles some of the historical tasks. Thus, quantifying task similarity is crucial to the effectiveness of a meta-learning based approach \cite{DBLP:journals/corr/abs-1810-03548}. 

Our work is most similar to the recent work by Zhao {\em et al.} \cite{DBLP:journals/corr/abs-2009-10606}  that developed the first unsupervised approach to outlier model selection. Their proposed {\sc MetaOD} quantifies task similarity based on \textit{meta-features}, which reflect general statistical properties of a dataset where the majority of the points are inliers. 
As such, two datasets with similar inlier distribution but different types of outliers may look similar w.r.t. meta-features, while different outlier models may be more effective in detecting different outlier types that they exhibit. In such scenarios, meta-feature similarity would be a poor indicator of model performance similarity.

In contrast to \metaod, we propose a \textit{performance-driven} similarity measure, where  two tasks are deemed more similar, the more the same set of models perform well/poorly on both tasks.
Arguably, ours is a \textit{direct} means to the end goal---only when task similarity is defined in this performance-based fashion would it be natural (and even gold standard) to choose a model for a new task that performs well on its neighbor tasks.

Our performance-driven similarity is a \textbf{rank-based} measure, called the weighted Kendall's tau rank correlation coefficient \cite{Shieh1998} ($\tau$ for short), which quantifies the similarity between the \textit{ordering of the candidate models by performance}. Formally, let $\bP_i \in \R^m$ depict the $i$-th row of $\bP$ containing the detection performances of models $\mM = \{M_1,\ldots, M_m\}$ on dataset $\mD_i$, and  let $\Delta_{j,j'}^{(i)} = \bP_{i,j}-\bP_{i,j^{\prime}}$ denote the difference between the performances of $M_j$ and $M_{j^{\prime}}$ on $\mD_i$.
Then, $\tau$ between two tasks $\mD_i$ and $\mD_{i^\prime}$ is defined as follows.
\begin{equation*}
     \tau^{(i,i^{\prime})}  = \tau(\bP_i,\bP_{i^{\prime}}) 
     = \frac{\sum_{j=1}^{m-1}\sum^m_{j^{\prime}=j+1} w_{j, j^{\prime}}^{(i, i^\prime)}}{\sum_{j=1}^{m-1}\sum^m_{j^{\prime}=j+1} |w_{j, j^{\prime}}^{(i, i^\prime)}|}
     \;, \quad \text{where}
\end{equation*}
\vspace{-0.1in}

\predisplaypenalty=150

\begin{minipage}{\columnwidth}
\begin{equation}
    w_{j, j^{\prime}}^{(i, i^\prime)}= 
\begin{cases}
        1  & \text{if } 
    \Delta_{j,j'}^{(i)} = \Delta_{j,j'}^{(i^\prime)}=0\\

    \Delta_{j,j'}^{(i)} \Big/ \Delta_{j,j'}^{(i^\prime)} 
    & \text{else if } 
    |\Delta_{j,j'}^{(i)}| \leq |\Delta_{j,j'}^{(i^\prime)}|     
    \\
    \Delta_{j,j'}^{(i^\prime)} \Big/ \Delta_{j,j'}^{(i)} 
    & \text{else; i.e. } 
    |\Delta_{j,j'}^{(i)}| > |\Delta_{j,j'}^{(i^\prime)}|\\
\end{cases}
\label{equ:kendall}
\end{equation}
\end{minipage}

Intuitively, $\tau$ quantifies the concordance/discordance between pairwise model performances, where $\tau$ becomes smaller both when the rank-orders are discordant (i.e., $w_{j,j'}$ is negative) and when they are concordant but the $\Delta$-differences are disproportionate (i.e., $w_{j,j'}$ is near-zero).
Put differently, two tasks are more similar when the models are ranked similarly, and also the perf. of each model are similar in value on both.

\subsection{\bf Internal Performance Measures (IPMs)}
\label{ssec:ipm}

To compute the similarity of a new task to historical meta-train tasks, we would need pairwise model performance differences on the new task. These model performances, of course, cannot be computed due to the lack of ground-truth labels.
In fact, having these at hand would obviate the model selection problem altogether.
Then, how can we really compute performance-driven task similarity?

The key idea is to learn a predictive model of ground-truth performance (which is available for meta-train tasks) from unsupervised indicators/features. The challenge is to identify such features that correlate with model performance. Notably, there exists a small literature on {internal evaluation} of outlier detection models \cite{goix2016evaluate,marques2020internal}, and recently also unsupervised model selection strategies for deep representation learning based on other internal measures \cite{DuanMSWBLH20,lin2020infogan}.
These are called \textit{internal} performance measures (IPMs) as they solely rely on the input samples, the outlier scores and/or the consensus between the candidate models. 
For example, consensus-based IPMs in principle associate closeness to an overall consensus of outlier scores/ranking with being a better model.
We refer to \S\ref{subsec:model_selection_od} for more details on specific IPMs.

An IPM is exactly designed to compare models without any ground truth labels. Then, the question is why do not we simply and directly use an IPM for model selection? The reasons are two-fold. First, effectiveness-wise, IPMs are weak/noisy signals of true performance. As a recent study showed, they enable only slightly better model selection than random  \cite{DBLP:journals/corr/abs-2104-01422}.
Distinctly, \method plugs the IPMs into a meta-learning framework, taking advantage of machine learning's ability to map weak signals onto desired ones. Related, it would not be clear \textit{which} IPM we should use for model selection (a ``chicken-egg'' scenario), provided several options. Notably, \method leverages all/any available IPMs as internal features in meta-training.
Second, computation-wise, 
we would need to train each and every candidate model on $\mDs$ to obtain its IPM and compare it to others,
leading to inhibitively high cost.
In contrast, \method 
employs meta-learning to identify similar historical tasks based on a small \textit{subset} of trained models on $\mDs$ while still being able to select among \textit{all} of the candidate models via their (ground-truth) performance on these neighbor tasks.

\subsection{\bf From IPMs to (True) Model Performance}
\label{ssec:performance_predictor}

At the core of \method's meta-training is {\em learning} to map IPMs onto ground-truth performance by the supervision from the meta-train database. In particular, \method learns a regressor that maps the IPMs from two models onto their performance difference. 

More formally, let $\phi(\cdot)$ denote the process of extracting various IPMs of model $M_j$ when trained on $\mD_i$, and $\bmij$ denote the corresponding vector of IPMs.
\method uses three IPMs; namely ModelCentrality (MC), HITS, and SELECT, as described in \cite{DBLP:journals/corr/abs-2104-01422}.
The regression function, named \textit{pairwise performance predictor}, maps the IPMs of any pair of models 
$M_j$ and $M_{j'}$ onto their performance difference on a dataset, i.e.
$f(\bmij, \bmijp) \mapsto \Delta_{j,j'}^{(i)}$. 
In implementation we use LightGBM \cite{DBLP:conf/nips/KeMFWCMYL17}, 
while it is flexible in choosing any other.
We design $f(\cdot)$ for pairwise prediction such that its output can be directly plugged into Eq. \eqref{equ:kendall} for computing task similarity. 
We find it important to remark that provided with $\phi(\cdot)$ and the trained $f(\cdot)$ at test time, measuring performance-driven task similarity via $\tau$ becomes possible without using any ground-truth labels.

For clarity of presentation, we defer a few implementation details to Appx. \S\ref{subsec:train_details}, where we describe  how to incrementally compute the IPMs as additional models are trained on the new input task at test time, as well as how to effectively train the regressor and use its predictions at test time.

\section{\method: Model Selection (Online)}
\label{sec:online_model_selection}
After the meta-training phase, \method is ready to admit a new task for model selection. Simply, it selects the highest performance model on meta-train tasks (or meta-tasks) that are very similar to the new task (\S\ref{ssec:neighbors}).
It identifies these similar meta-tasks iteratively by refining the similarity estimates adaptively (\S\ref{ssec:modelsubset}).

\vspace{-0.05in}
\subsection{\bf Model Selection via Similar Meta-tasks}
\label{ssec:neighbors}

Given a new task $\mDs$, we aim to identify its similar meta-train tasks (referred as the neighbor set $\mN \subset \mDt$). By the principle of meta-learning, the model(s) that outperform on the neighbor set is likely to outperform on the new task as well. Consequently, we could output the model with the largest average performance on the neighbor set as the selected model for the new task, that is, 
\begin{equation}
\vspace{-0.05in}
    \argmax_{M_j \in \mM} \; \frac{1}{|\mN|}\sum_{\mD_i \in \mN} \bP_{i, j} \;.
    \label{equ:selected_model}
\end{equation}

If the model performances $\bP_{\text{test}}$ were available for $\mDs$, we could iterate over the meta-train database to measure task similarity via Kendall's tau in Eq. (\ref{equ:kendall}), 
and pick $\mN$ to be the top $t$ most similar meta-train tasks, as shown in Eq. (\ref{equ:neighbor_set_orcale}). Here $t$ denotes the size of the neighbor set (i.e. $|\mN|=t$), which can be chosen by cross-validation on the meta-train database (see Appx. \S \ref{subsubsec:hyperparameters} for details).
\begin{equation}
    \mN := \underset{\scriptscriptstyle i=1 \ldots n}{\text{top-}t}\; \tau^{(\text{test}, i)} =
    \underset{\scriptscriptstyle i=1 \ldots n}{\text{top-}t} \; \tau(\bP_{\text{test}}, \bP_i)
    \label{equ:neighbor_set_orcale}
\end{equation}

However, we cannot directly identify
$\mN$ 
due to the lack of ground-truth labels and thus evaluations $\bP_{\text{test}}$ on the new task. 
Note the calculation of Kendall's tau in Eq. (\ref{equ:kendall}) only depends on pairwise performance gaps (i.e. $\Delta$-differences) to measure concordance/discordance, where the trained regressor $f(\cdot)$
for pairwise performance gap prediction 
comes into play.
By plugging the \textbf{predicted pairwise gaps} (i.e.  $\widehat{\Delta}$-differences) of the new task into Eq. (\ref{equ:kendall}), we could therefore estimate its 
neighbor set $\mN$  even when $\bP_{\text{test}}$ is inaccessible.

Specifically, for the new task $\mDs$, we first get its outlier scores  $\mO_{\text{test}} = \mM(\mDs)$ and build the IPMs  $\mathbf{m}_{\text{test}}=\phi(\mO_{\text{test}})$ across candidate models.Note that We slightly abuse notation here and use $\mathbf{m}_{\text{test}}$ to depict the IPM vectors for all models, which is in fact a matrix. We could then predict the performance gap of any pair of models for $\mDs$ using the regressor $f(\cdot)$ by
\begin{equation}
    \widehat{\Delta}_{j,j^{\prime}}^{(\text{test})} := f(\mathbf{m}_{\text{test}, j}, \mathbf{m}_{\text{test}, j^{\prime}}) \approx
    \Delta_{j,j^{\prime}}^{(\text{test})}\;,
    \label{eq:predicted_gap}
\end{equation}

where $j=1 \ldots m$ and $ j < j^{\prime}$.  
The estimated Kendall-tau similarity ($\widehat{\tau}$) between the new task and meta-train tasks can be calculated using the \textit{predicted} pairwise performance gaps on the new task $\widehat{\Delta}^{\text{test}}$ in Eq. (\ref{eq:predicted_gap}) and the \textit{actual} performance gaps on meta-train tasks $\Delta^{\text{train}}$, 
where e.g., we denote by $\widehat{\tau}^{\;(\text{test},i)}$ 
the estimated Kendall-tau similarity between $\mDs$ and the $i$-th meta-train dataset.
 
Then, by 
 plugging the estimated task similarities into Eq. (\ref{equ:neighbor_set_orcale}), we obtain the neighbor set $\mN$ of the new task as the top-$t$ meta-train datasets with the highest estimated Kendall-tau similarity 
\textit{without relying on ground-truth labels}, i.e.,
\begin{equation}
    \mN :=
    \underset{\scriptscriptstyle i=1...n}{\text{top-}t}\;\; \widehat{\tau}^{\;(\text{test},i)}
    \approx
    \underset{\scriptscriptstyle i=1...n}{\text{top-}t}\;\; \tau^{(\text{test}, i)} \;.
    \label{eq:neighbor_set_estimate}
\end{equation}

\subsection{\bf Unsupervised Adaptive Search}
\label{ssec:modelsubset}

\subsubsection{\bf Motivation and Initialization}
\label{sssec:init}

Quantifying the task similarity by Kendall-tau using the full model set $\mM = \{M_1,\ldots, M_m\}$ (based on all 
${m \choose 2}$
pairs of performance gaps) 
incurs high computational cost, as it involves 
model fitting to get outlier scores 
and extracting corresponding IPMs
for {\em all} candidate models. We therefore propose to only measure  task similarity based on a \textit{subset} of the models, denoted $\mM_s \subset \mM$ for model subset. 
Initially, 
$\mM_s$ can be set to a small random subset of $\mM$, while a more careful initialization strategy may facilitate better similarity measurement. In \method, we design a coverage-maximization strategy for initializing $\mM_s$,  details of which are described in Appx. \S \ref{subsubsec:model_init}. The ablation in \S \ref{subsubsec:ablation_init} shows it is significantly better than random initialization.

\subsubsection
{\bf Iteration: Adaptively Expanding Model Subset}
\label{subsubsec:adptive_search}
The initial $\mM_s$ may not be sufficient to capture a complete picture of task similarity. 
Therefore, \method expands the model subset iteratively to refine the task similarity estimates and thereby obtain increasingly better estimates of the most similar datasets to $\mDs$.

Assume for now that an objective criterion exists for choosing the next model to be included in $\mM_s$, then, the adaptive search proceeds as follows. In each iteration, we update the neighbor set $\mN$ by the Kendall-tau similarities in Eq. \eqref{equ:kendall} computed based on the model subset $\mM_s$. Then, we quantify the value of each candidate model that is not already in $\mM_s$ against the objective criterion and expand the model subset with the model $M^*$ 
having the maximum value, i.e. $\mM_s:= \mM_s \cup M^*$.
In this way, we only need to 
fit on $\mDs$
(and get the corresponding IPMs)  
for the newly added model $M^*$ per iteration. 
To reduce the overall computational cost, the goal is to accurately identify highly similar neighbors $\mN$ based on as \textit{few} models trained on $\mDs$ as possible.
It is important to note, however, that even if we train only a small subset of the models on the new task, upon identifying $\mN$, we select a model from among {\em all}  candidate models using Eq. \eqref{equ:selected_model}.

What objective criterion is suitable for iteratively choosing the next model to be included in the model subset $\mM_s$?
We argue that the added model should meet two criteria, \textit{uncertainty} and \textit{quality}: 
\begin{criterion}[Uncertainty]
The performance (rank) of the added model should \textit{\it vary} across
the current neighbor set $\mN$.
\label{criterion:Variability}
\end{criterion}

\begin{criterion}[Quality]
The added model should \textit{\it outperform} on the current neighbor set $\mN$. 
\label{criterion:Quality}
\end{criterion}

Notably, without \textit{uncertainty}, $\mM_s$ would end up choosing a group of similar models without enough representation of the full model space, which inhibits finding truly similar meta-tasks. A model with high performance variance over $\mN$ indicates the datasets within the neighbor set exhibit disagreement (i.e. neighbors are not as similar among themselves), and including the model unlocks the opportunity to find truly similar meta-tasks. On the other hand, the \textit{quality} criterion emphasizes that the added model should be a well-performing model on the neighbor set, as model selection mainly concerns ``top models''. As such, we aim to identify the neighbor set based on task similarity regarding the well-performing models, whereas performance similarity based on underperforming models does not contribute much to the main goal of (top) model selection.

Now it is easy to see the \textbf{trade-off} between \textit{uncertainty} and \textit{quality}---the former emphasizes the model's performance variation among the neighbor set while the latter expects high performance over all.
This is akin to the explore-exploit trade-off; \textit{uncertainty} drives exploration (for better neighbors) while \textit{quality} drives exploitation (by promptly pinning a top model).

How can we quantify the \textit{uncertainty} and \textit{quality} of a candidate model (say, $M_j$)? Naturally,
they can be respectively measured as the \textit{variance} of the model's performance, $\sigma^2_{j}=\sigma^2(M_j| \mN)$, and its \textit{average} performance, $\mu_j = \mu(M_j| \mN)$, given the neighbor set. 
Thus, the simplest objective criterion that considers both \textit{uncertainty} and \textit{quality} criteria can be defined as the sum of the two: 
\vspace{-0.1in}
\begin{equation}
    \argmax_{M_j \in \mM \setminus \mM_s}\;\;
   \underbracket{\sigma^2(M_j| \mN))}_\textrm{Uncertainty}+
   \underbracket{\mu(M_j|\mN))}_\textrm{Quality}
    \label{equ:simple_goal}
\end{equation}
\vspace{-0.1in}

Can we define a better objective that automatically balances the trade-off between the two criteria? At this stage, 
we can recognize a connection 
to Sequential Model-based Bayesian Optimization (SMBO) \cite{DBLP:journals/jgo/JonesSW98}.
As discussed in \S \ref{subsec:model_select_hyper_tuning}, 
SMBO is a state-of-the-art paradigm for solving sequential problems like hyperparameter optimization in supervised settings \cite{DBLP:conf/nips/SnoekLA12}.
As an iterative method, it relies on what-is-called an \textit{acquisition function} $a(\cdot)$ that quantifies the utility of a candidate hyperparameter configuration (HPC) for the next evaluation \cite{DBLP:journals/jgo/JonesSW98}. 
In fact, as with the uncertainty/exploration and quality/exploitation trade-off, $a(\cdot)$ typically aims to balance between picking an HPC from the unexplored regions of the hyperparameter space and one with high estimated accuracy.

To capitalize on this connection, \method 
leverages the prominent acquisition function in SMBO called Expected (positive) Improvement (EI) \cite{DBLP:journals/jgo/JonesSW98} as our objective criterion to automatically balance the uncertainty-quality trade-off. In our setting, EI measures the expected improvement of including a candidate model into the model subset. The high EI value of a candidate model means that it has large performance variation and also high performance over the neighbor set $\mN$. 
Moreover, one of the nice properties of EI is it has a closed-form expression under the Gaussian assumption, where the EI of a candidate model is defined as:
\begin{equation}
\label{eq:EI}
    EI(M_j|\mN):=\sigma_j \cdot [u_j\cdot \Phi(u_j)+\varphi(u_j)], \quad \text{where}
    \vspace{-0.1in}
\end{equation}
\begin{equation*}
 u_j= 
\begin{cases}
    \frac{\mu_j-\mu_s^{*}}{\sigma_j}  & \text{if } \sigma_j > 0\;; \quad \textrm{and } \quad
    \bigg\{ 0 \quad \text{if } \sigma_j = 0
\end{cases}\;.
\end{equation*}

In the above, $\Phi$ and $\varphi$ respectively denote the cumulative distribution and the probability density functions of standard Normal distribution, $\mu_j$ and $\sigma_j$ are the mean and the standard deviation of model $M_j$ across the neighbor set $\mN$, and $\mu_s^{*}$ is the maximum value of the average model performance of $\mM_s$ over $\mN$, i.e.,  

\vspace{-0.1in}
\begin{equation*}
\mu_s^{*}= \underset{\scriptsize M_h \in \mM_S}{\max} \;\; \frac{1}{|\mN|}\sum_{\mD_i \in \mN} \bP_{i, h} \;.
\end{equation*}
\vspace{-0.1in}

With the EI-based objective in Eq. (\ref{eq:EI}), we include the highest EI candidate model to the model subset per iteration:

\vspace{-0.1in}
\begin{equation}
    \mM_s := \mM_s \cup \argmax_{M_j \in \mM \setminus \mM_s}
    EI(M_j|\mN)\;.
    \label{equ:goal}
\end{equation}
\vspace{-0.1in}

To sum up, \method alternates between ($i$) updating the neighbor set $\mN$ using Eq. \eqref{eq:neighbor_set_estimate} based on the (expanded) model subset $\mM_s$, and ($ii$) expanding $\mM_s$ using Eq. \eqref{equ:goal} based on the (updated) $\mN$, until converged or termination criteria are met.

\subsubsection{\bf Convergence and Termination Criteria}
\label{subsubsec:convergence}

\method can operate under two different practical settings: (1) {\em hands-off}: there is no time budget and (2) {\em hands-on}: the user has time constraints and the algorithm is to output a selected model whenever prompted.

\textit{Hands-off}: When there is \textit{no} time budget, \method stops when the neighbor set $\mN$ stays unchanged in $p$ consecutive iterations, indicating that the identified similar meta-tasks have stabilized.
We refer to parameter $p$ as ``patience'', as a larger $p$ 
requires more iterations to converge. In the extreme case (when $p$ is set to a very large value), the algorithm would stop when all the models are added to the model subset, i.e. $\mM_s=\mM$. In this case, \method measures task similarity based on all models, falling back to the original setting (\S \ref{ssec:neighbors}) \textit{without} the adaptive expansion. $p$ can be decided by cross-validation; see details in Appx. \S \ref{subsubsec:hyperparameters}.

\textit{Hands-on}: When there is a time budget $b$ (iterations) to accommodate, \method stops the adaptive search when whichever one of two conditions occurs earlier: time budget is up, or patience criterion above is met. Note that the time budget need not be known to \method apriori, for it 
is an ``\textit{any-time} algorithm'':
at any time the user prompts it during its course, 
it can always output a selected model as the one with the highest average performance on the neighbor set at the current iteration, based on Eq. (\ref{equ:selected_model}).

\subsection{Computational Complexity}
\label{subsec:complexity}
Suppose that a task has $r$ samples and $d$ features on average, and the score computation of an OD model takes $C_{\text{train}}(r,d)$.


\begin{lemma}[Meta-training]{The computational complexity of \method's meta-training phase (offline) is $O(nmr+nm^2)$. 
}\label{lemma:offlinecomplexity}
\end{lemma}
\begin{lemma}[Model selection]{The computational complexity of \method's model selection phase (online) is $O[b(bn+mt+C_{\text{train}}(r,d))]$, for budget $b$ and neighbor count $t$.}\label{lemma:onlinecomplexity}
\end{lemma}

\noindent See details in Appx. \ref{sec:complexity}. Note that  the quadratic
$m^2$ term in offline training is for measuring \textit{pairwise} performance-driven task similarity, where $m$ is  small (e.g., 297 in this study). In the online phase, the complexity is linear in both the number of meta-train datasets ($n$) and that of candidate models ($m$).

\vspace{-0.1in}

\vspace{0.05in}
\section{Experiments}
\label{sec:experiments}
\subsection{Experiment Setting}
\label{ssec:setting}

\noindent {\bf Model Set.} We configure 8 leading OD algorithms with various different settings of their associated hyperparameters to compose the model set $\mM$ with 297 models (based on MetaOD \cite{DBLP:journals/corr/abs-2009-10606}; the only diff. is we set n\_neighbors of ABOD to $[3,5,10,15,20,25,50]$ for faster experimentation).
We evaluate \method in two testbeds introduced below, with 39 and 30 datasets, respectively.
For each testbed, we first generate the outlier scores of each model in $\mM$ on each dataset, and then record the historical performance matrix $\bP$. For models with built-in randomness, e.g., iForest and LODA with random feature splits, we run 5 random trials and record the average. 
All OD models are built using the PyOD library \cite{DBLP:journals/jmlr/ZhaoNL19} on an Intel i7-9700 @3.00 GHz, 64GB RAM, 8-core workstation. 

\begin{figure}[!ht]
\centering
\vspace{-0.1in}
	\includegraphics[width=0.85\linewidth]{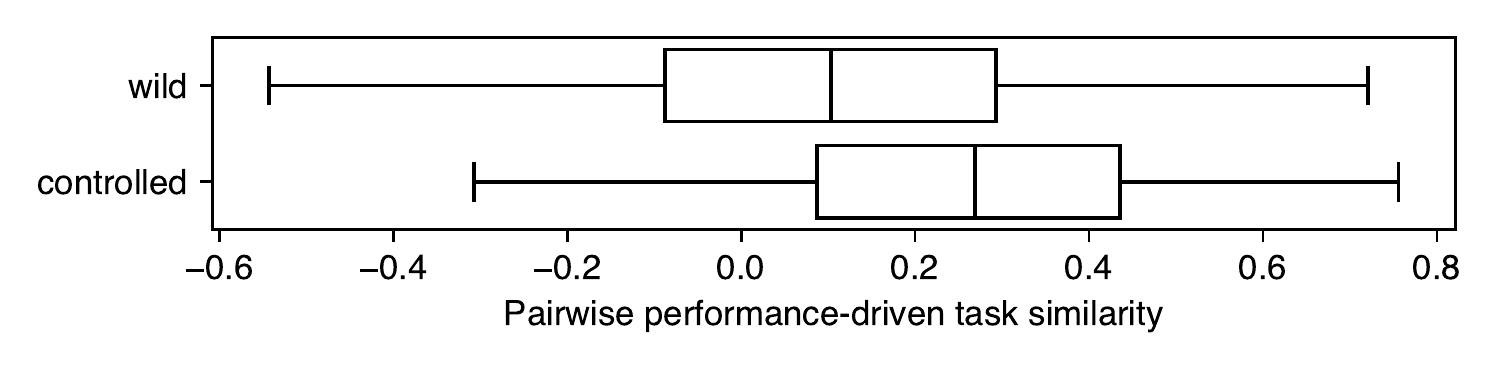}
	\vspace{-0.15in}
	\caption{ Pairwise performance-driven task similarity in the wild (med.=0.1035; lower similarity) and controlled (med.=0.2688; higher similarity) testbed. 
	}
	\label{fig:testbed_similarity}
	\vspace{-0.1in}
\end{figure}

\noindent
\textbf{Testbeds}. The key mechanism of meta-learning is to leverage the prior knowledge from truly similar tasks---where OD models perform similarly on both tasks. 
We create two testbeds with varying performance-driven task similarities (see Fig. \ref{fig:testbed_similarity}).

\noindent (1) \textbf{Wild testbed} contains 39 independent datasets from 2 public OD repository 
(i.e., ODDS \cite{Rayana:2016} and DAMI \cite{Campos2016})
which simulates real-world use cases. 

\noindent (2) \textbf{Controlled testbed} contains 30 datasets generated from 10 independent ``mothersets'' (from the Wild testbed), where 3 types of outliers (global, local, clustered) are injected into each motherset. 
As such, higher performance-driven task similarities are expected in the datasets with the same type of injected outliers (but from different mothersets), while their meta-feature similarities are low due to the independence of the mothersets. This testbed helps understand 
(\textit{i}) if meta-feature methods can work when  meta-feature similarity 
disagree with
performance-driven similarity and 
(\textit{ii}) to what extent the level of task similarity affects the performance of \method.

\noindent{\bf Baselines.} We include 13 baselines for comparison. As shown in Table \ref{table:baseline}, they can be categorized by (i) whether it selects a model; (ii) whether it is based on meta-learning; and (iii) whether it relies on meta-features. We use all 13 baselines in the wild testbed, and the 5 meta-feature-based methods in the controlled testbed as it is built to contrast performance- vs. meta-feature-based task similarities.

\setlength\tabcolsep{1.5 pt}

\begin{table}[t]
\vspace{-0.1in}
\centering
	\caption{13 baselines for comparison with categorization by (first row) whether it is a model selection method (second row) whether it uses meta-learning and (third row) whether it relies on meta-features (last row).} 
\vspace{-0.1in}
\scalebox{0.65}{
\begin{tabular}{l|cccccc|ccccccc}
\toprule
\textbf{Category}      & \textbf{iForest} & \textbf{LOF} & \textbf{ME} & \textbf{MC} & \textbf{SELECT} & \textbf{HITS} & \textbf{GB} & \textbf{ISAC} & \textbf{AS} & \textbf{ALORS} & \textbf{MetaOD} & \textbf{SS} & \textbf{IPM\_SS} \\
\midrule
\textbf{model selection} &                 &           &          &        $\bullet$  &          $\bullet$    &           $\bullet$ &     $\bullet$     & $\bullet$           & $\bullet$         & $\bullet$            & $\bullet$             & $\bullet$         &   $\bullet$     \\ 
\textbf{meta-learning} &               &           &          &          &              &            & $\bullet$         & $\bullet$           & $\bullet$         & $\bullet$            & $\bullet$             & $\bullet$         & $\bullet$              \\
\textbf{meta-features}  &               &           &          &          &              &            &          & $\bullet$           & $\bullet$         & $\bullet$            & $\bullet$             & $\bullet$         &        \\     
\bottomrule
\end{tabular}
}
	\label{table:baseline} 
	\vspace{-0.3in}
\end{table}
\setlength\tabcolsep{6 pt}

Briefly, the baselines are organized as: 
(\textit{i}) \textbf{\textit{no model selection}}: directly/always use 
the same popular model \textbf{(1) iForest} \cite{liu2008isolation} or \textbf{(2) LOF} \cite{breunig2000lof}, or the ensemble of all models \textbf{(3) Mega Ensemble (ME)}; 
(\textit{ii}) \textbf{\textit{direct use of IPMs for model selection}}: \textbf{(4) MC} \cite{DBLP:journals/corr/abs-2104-01422}, \textbf{(5) SELECT} \cite{DBLP:journals/corr/abs-2104-01422}, and \textbf{(6) HITS} \cite{DBLP:journals/corr/abs-2104-01422}; and 
(\textit{iii}) \textbf{\textit{meta-learning based methods}}:  \textbf{(7) Global Best (GB)} selects the best performing model on meta-train database on average, \textbf{(8) ISAC} ~\cite{conf/ecai/KadiogluMST10},
\textbf{(9) ARGOSMART (AS)} ~\cite{nikolic2013simple}, \textbf{(10) ALORS} \cite{journals/ai/MisirS17}, \textbf{(11) MetaOD} ~\cite{DBLP:journals/corr/abs-2009-10606} is the SOTA method, \textbf{(12) Supervised Surrogates (SS)} \cite{xu2012satzilla2012} directly regresses meta-features to performance 
$\bP$, and \textbf{(13) IPM-based Supervised Surrogates (IPM\_SS)} is a variant of SS but uses IPMs other than meta-features. Baselines (8)-(12) use meta-features.

\noindent {\bf Evaluation.} In both testbeds, we use leave-one-out cross-validation (LOOCV) to split the meta-train/test. Each time we use one dataset as the input task, and the remaining datasets as meta-train. Meanwhile, we use cross-validation to decide the size of the neighbor set $\mN$ for each input task, and use a fixed time budget $b=50$ as the convergence criteria.
We use the area under the precision-recall curve (Average Precision or AP) as the performance measure, while it can be substituted with any other measures, e.g., the area under the receiver operating characteristic curve (ROC). Since the raw performance like AP is not comparable across datasets with varying magnitude, we report the AP-rank of a selected model, ranging from 1 (the best) to 297 (the worst)---thus smaller the better.
To compare two methods, we use the paired Wilcoxon signed rank test across all datasets in the testbed (significance level $p < 0.05$). 

\subsection{Experiment Results}

\begin{figure}[!htb]
\centering
	\includegraphics[width=0.9\linewidth]{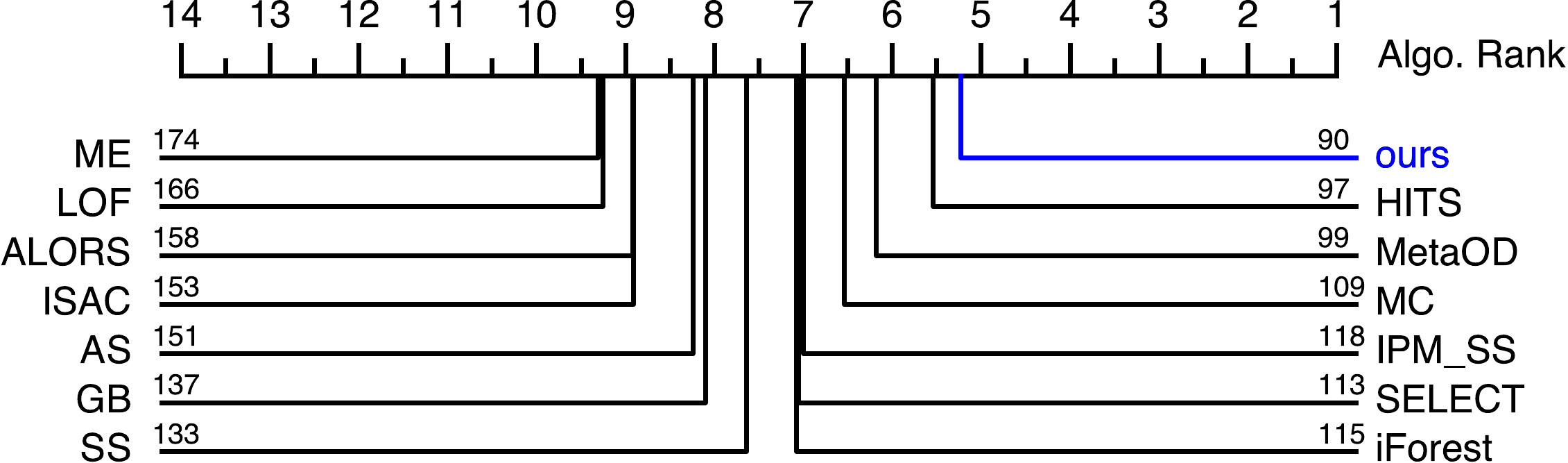}
	\vspace{-0.15in}
	\caption{Comparison of avg. rank (lower is better) of algorithms w.r.t. performance across datasets in the wild testbed. \method outperforms all w/ the lowest avg. rank.
	Numbers on each line are the avg. AP-rank (lower is better) of the employed model (selected or otherwise) 
	by each method. 
	}
	\label{fig:cd_plot_39}
	\vspace{-0.1in}
\end{figure} 

\subsubsection{\bf Results on Wild Testbed}
As shown in
 Fig. \ref{fig:cd_plot_39}, 
 \textbf{\method outperforms all baseline algorithms with the lowest avg. rank} ``in the wild''. Furthermore, Table \ref{table:pairs} (left) shows that \method is the only algorithm that is not significantly different from the 55-$th$ best model. In other words, \method can consistently choose the top 18.5\% model from a large pool of 297 models. Moreover, \method is significantly better than the no model selection baselines, LOF, iForest, ME, and other meta-learning baselines including GB, ISAC, ALORS, SS, and IPM\_SS.
 For other baselines, p-value remains low although not significant at 0.05.

\begin{table}[!t]
\scriptsize
\caption{
Pairwise statistical tests between ELECT and baselines by Wilcoxon signed rank test (statistically better method at $p$$<$$0.05$ in \textbf{bold},  both in \textbf{bold} if no difference). 
In wild testbed (left), ELECT is the only approach with no difference from the 55-$th$ best model. In  controlled testbed (right), compared with meta-feature based baselines, \method is the only method with no  difference from the 32-$th$ best model, and statistically better than all baselines.
}
\vspace{-0.1in}
\label{table:pairs}
\scalebox{1}{
\begin{tabular}{ll}
\centering
\hspace{-0.1in}
\begin{tabular}{ll|l}
\toprule
\textbf{Ours} & \textbf{Baseline} & \textbf{p-value} \\
\midrule
\textbf{ELECT}        & \textbf{55-th Best}         & 0.0541        \\
\midrule
\textbf{ELECT}        & iForest               & 0.0008                 \\
\textbf{ELECT}        & LOF           & 0.0004                  \\
\textbf{ELECT}        & ME                & 0.0188                \\
\textbf{ELECT}        & \textbf{MC}                & 0.137                 \\
\textbf{ELECT}        & SELECT              & 0.0484                \\
\textbf{ELECT}        & \textbf{HITS}                & 0.2142        \\
\textbf{ELECT}        & GB                & 0        \\
\textbf{ELECT}        & ISAC             & 0                 \\
\textbf{ELECT}        & AS               & 0.0147        \\
\textbf{ELECT}        & ALORS               & 0.0002        \\
\textbf{ELECT}        & \textbf{MetaOD}               & 0.2766        \\
\textbf{ELECT}        & SS               & 0.0128        \\
\textbf{ELECT}        & IPM\_SS               & 0.0019        \\
\bottomrule
\end{tabular}
     & 
\begin{tabular}{ll|l}
\toprule
\textbf{Ours} & \textbf{Baseline} & \textbf{p-value} \\
\midrule
\textbf{ELECT}        & \textbf{32-th Best}         & 0.0631        \\
\midrule
ELECT       & iForest               & N/A                 \\
ELECT       & LOF           & N/A                  \\
ELECT       & ME                & N/A                \\
ELECT       & MC                & N/A                 \\
ELECT       & SELECT              & N/A                \\
ELECT        & HITS                & N/A        \\
ELECT        & GB                & N/A        \\
\textbf{ELECT}        & ISAC             & 0.0012                 \\
\textbf{ELECT}        & AS               & 0.003        \\
\textbf{ELECT}        & ALORS               & 0.0001        \\
\textbf{ELECT}        & MetaOD               & 0.0016        \\
\textbf{ELECT}        & SS               & 0.0007        \\
ELECT       & IPM\_SS               & N/A        \\
\bottomrule
\end{tabular}
\end{tabular}}
\vspace{-0.2in}
\end{table}

\begin{figure}[!t]
\centering
	\includegraphics[width=0.85\linewidth]{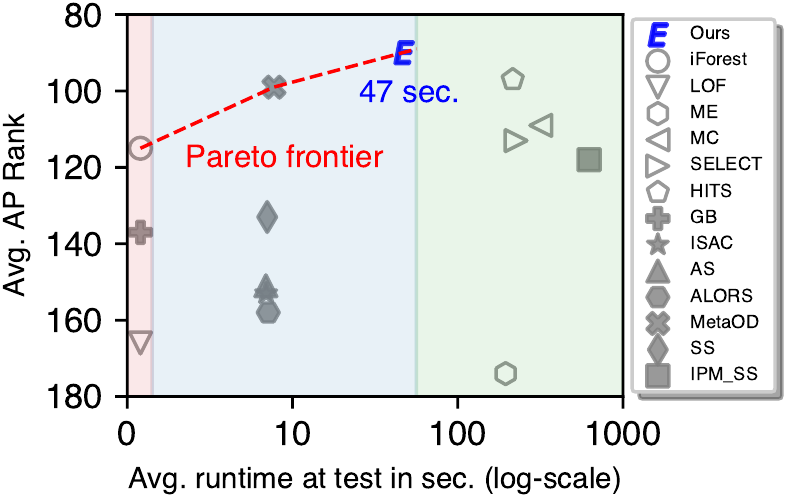}
	\vspace{-0.125in}
	\caption{Avg. running time (log-scale) vs.  avg. model AP-rank.
	 Meta-learning methods depicted w/ solid markers.  Based on  runtime,  methods are categorized as i) super-fast (in \crule[alicered]{0.2cm}{0.2cm}), ii) fast (in \crule[aliceblue]{0.2cm}{0.2cm}), and iii) slow  (in \crule[alicegreen]{0.2cm}{0.2cm}). Pareto frontier (red dashed line) shows the best method under different time budgets. 
	 \method outperforms all with small time consumption (on avg. below 1 min. per task).
	}
	\label{fig:pareto_frontier}
	\vspace{-0.25in}
\end{figure}

\noindent {\textbf{\method achieves the best performance with small ($<$$1$ min.) overhead}}. 
Fig. \ref{fig:pareto_frontier} shows the running time of the methods versus the avg. AP-rank of the employed model. Based on the avg. selection time per dataset, the methods can be categorized as i) super-fast methods that take less than 1 sec. (red zone); ii) {fast} methods that take less than 1 min. (blue zone); and iii) slow methods that use up to 10 min.s (green zone). 
{Super-fast} methods either directly employ a model (iForest, LOF) or simply report the historical best model (GB) with limited performance,
showing the necessity for  more effective meta-learning. 
In time-critical applications, employing iForest is a reasonable choice and it is indeed on the Pareto frontier of the time-performance trade-off. In the {fast} group, both \metaod and \method are also on the Pareto frontier, showing the premise of effective meta-learning and the additional benefit of \method. 
{Specifically, \method (avg. AP-rank$=$$90$) brings 10\% performance improvement over \metaod (avg. AP-rank$=$$99$), while being fast (avg. time$=$$47.10$s)}.
For the slow group, in contrast, the higher runtime for model and IPM building does not yield improved performance  over the {fast} methods.

\noindent \textbf{IPMs do carry useful signals, and \method can leverage them more effectively}. In fact, IPM-based MC, SELECT, and HITS rank high among all methods
(Fig. \ref{fig:cd_plot_39}), showing their potential in model selection.
However, using IPMs directly for model selection incurs a large overhead in building all OD models and IPMs themselves at selection time, and therefore all of them fall in the slow group as shown in Fig. \ref{fig:pareto_frontier}.
Building on top of IPMs, \method shows superior results by ``juicing out'' useful information from IPMs via meta-learning, as well as reducing the runtime via sequential learning to prevent excessive model building.

\subsubsection{\bf Results on Controlled Testbed}
\label{sssec:controlled}
\hfill

\noindent \textbf{Setup details.} Meta-learning facilitates model selection for a new task by leveraging the prior knowledge from its truly similar meta-train tasks---where OD models \textit{perform} similarly. The controlled testbed is built to create the scenario where there exist meta-train 
tasks with \textit{high performance-driven similarity} but \textit{low meta-feature similarity}, and vice versa. As meta-feature methods assume a high correlation between meta-feature similarity and task-performance similarity, they are likely to do poorly (i.e. select poor models) in this testbed as the assumption is violated.

To create such a setting, we randomly select 10 independent ``mothersets'' from the wild testbed, 
and inject one of 3 types of synthetic outliers (global, local, and clustered)  by following \cite{steinbuss2021benchmarking}, resulting in 30 datasets. Intuitively, different OD models are good at successfully detecting different types of outliers, irrespective of the underlying motherset. Therefore, we expect the datasets from different mothersets but with the same type of injected outliers to have high performance-driven task similarity yet low meta-feature similarity, while the datasets from the same mothersets but with different types of injected outliers to be vice versa. 

\begin{figure}[H]
\centering
\vspace{-0.1in}
	\includegraphics[width=0.85\linewidth]{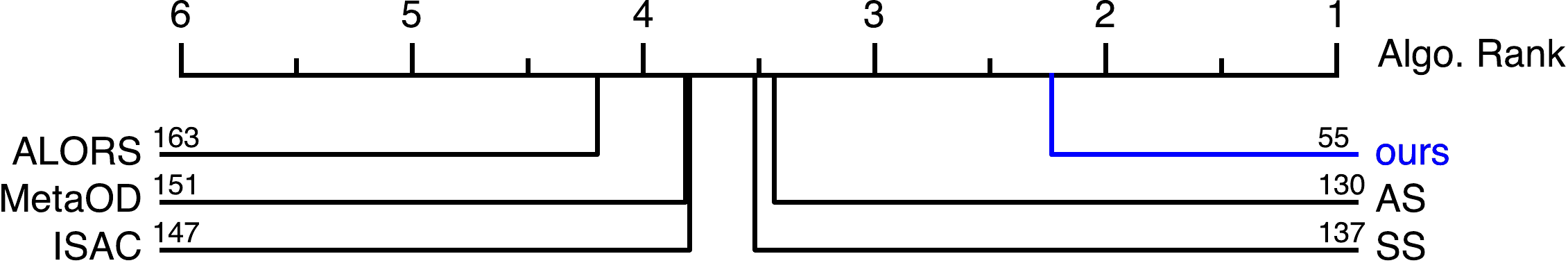}
	\vspace{-0.15in}
	\caption{Comparison of avg. rank (lower is better) of algorithms w.r.t. performance across datasets in the controlled. 
	\method outperforms all baselines.}
	\label{fig:cd_plot_db2}
	\vspace{-0.1in}
\end{figure} 

\noindent \textbf{Results.} As shown in Fig. \ref{fig:cd_plot_db2}, \textbf{\method is superior to all meta-feature-based baselines}. Note that its avg. AP-rank is 55, while the avg. AP-rank of meta-feature baselines are all above 130. 
All these differences are significant at $p<0.005$ as shown in Table \ref{table:pairs} (right). 
These results together show that the performance of meta-feature baselines suffers when performance-driven similarity is not correlated with meta-feature similarity. Distinctly, \method does not rely on meta-features and instead directly focuses on performance-driven similarity, leading to superior selection results.

\noindent \textbf{As a meta-learning method, \method achieves better results with higher task similarity to meta-train}. As shown in Fig. \ref{fig:testbed_similarity}, the controlled testbed has higher task similarity than the wild. In Table \ref{table:pairs} (right)  \method is the only method that shows no statistical difference from the 32-$th$ best model, suggesting it can choose the top 11\% model from $\mM$, an improvement from the top 18.5\% in the wild testbed. The selected models' avg. AP-rank reflects the same---55 vs. 90 in the controlled and wild testbed, respectively.

\vspace{-0.05in}
\subsection{Case Study}
\label{subsec:case_study}

It is interesting to trace how \method works over iterations on a given dataset, 
as shown in
Table \ref{table:case_study} 
for the Waveform dataset from the wild testbed. 
First we track 
the changes in the
neighbor set: 
col. 2 reports the avg. similarity between Waveform and the identified neighbor set $\mN$, and col. 3 shows the number of  ground-truth top 5 most similar meta-train tasks in $\mN$. 
\method gradually identifies both more similar and more of the top 5 meta-train tasks. From 1-$st$ to the 50-$th$ iteration, the avg. task similarity improves from 0.2276 to 0.3504.,
and the number of identified top 5 neighbors increases from 1 to 4. Note that out of the 38 meta-train tasks, only 7 of them have higher similarity than 0.3504.
By identifying more and more similar meta-train tasks, 
\method
gradually converges to kNN models as given in col. 4, which is indeed the best algorithm family for Waveform. Moreover, col. 5 shows that \method successfully identifies better models---the selected model's AP-rank decreases 
from 77 to 38. 
\vspace{-0.1in}

\begin{table}[!htb]
\centering
	\caption{Trace of \method on Waveform dataset. Over iterations (col. 1), \method gradually identifies more similar meta-train tasks with increasing avg. similarity to Waveform (col. 2), more ground-truth top 5 neighbors (col. 3), and a better selected model with lower rank (col. 4). Best performing algorithm family on Waveform is kNN, which \method successfully identifies during its adaptive search.} 
	\vspace{-0.1in}
	\footnotesize
	
\scalebox{0.8}{
	\begin{tabular}{c|cc|cc}
	\toprule
\textbf{Iter.} & \makecell{\textbf{Avg. } \\ \textbf{ Sim.}} & \makecell{\textbf{\# Matched} \\ \textbf{neighbors}}  & \textbf{The selected model}     & \makecell{\textbf{The selected } \\ \textbf{ model rank}} \\
\midrule
1                     & 0.2776                                   & 1                           & ('LOF', ('manhattan', 5))  & 77                         \\
2                     & 0.2876                                   & 1                           & ('COF', 10)                & 70                         \\
$\ldots$ &$\ldots$&$\ldots$&$\ldots$&$\ldots$\\
11                    & 0.3304                                   & 2                           & ('LOF', ('manhattan', 10)) & 76                         \\
12                    & 0.3493                                   & 3                           & ('LOF', ('euclidean', 20)) & 71.5                       \\
$\ldots$ &$\ldots$&$\ldots$&$\ldots$&$\ldots$\\
21                    & 0.4351                                   & 4                           & ('kNN', ('mean', 5))       & 50                         \\
22                    & 0.4351                                   & 4                           & ('kNN', ('mean', 5))       & 50                         \\
$\ldots$ &$\ldots$&$\ldots$&$\ldots$&$\ldots$\\
31                    & 0.2960                                   & 4                           & ('kNN', ('mean', 15))      & 38                         \\
32                    & 0.4351                                   & 4                           & ('kNN', ('mean', 5))       & 50                         \\
$\ldots$ &$\ldots$&$\ldots$&$\ldots$&$\ldots$\\
49                    & 0.3504                                   & 4                           & ('kNN', ('mean', 15))      & 38                         \\
50                    & 0.3504                                   & 4                           & ('kNN', ('mean', 15))      & 38      \\    \bottomrule          
\end{tabular}
}
	\label{table:case_study} 
	\vspace{-0.15in}
\end{table}

\vspace{-0.05in}
\subsection{Ablation Studies and Other Analysis}
\subsubsection{Model initialization}
\label{subsubsec:ablation_init}
\begin{figure}[H]
\centering
	\includegraphics[width=0.85\linewidth]{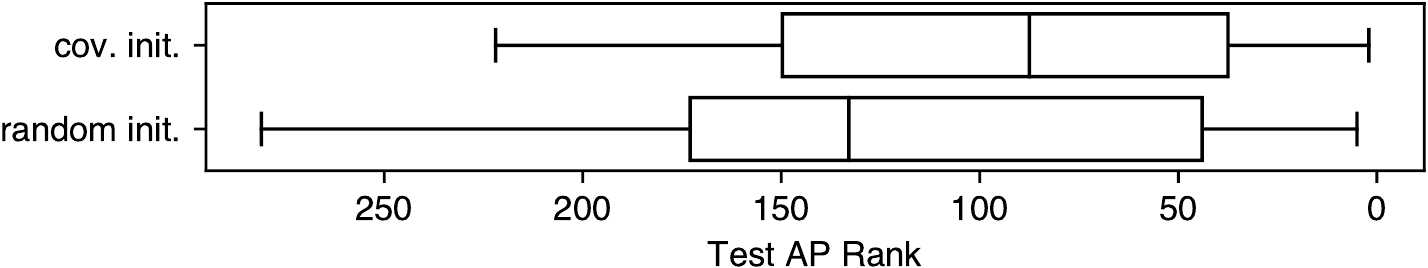}
	\vspace{-0.1in}
	\caption{Ablation of coverage init. (med.=87.5) vs. random init. (med.=133).
	}
	\label{fig:ablation_random}
	\vspace{-0.1in}
\end{figure}

\method uses proposed coverage-driven initialization of the model subset $\mM_s$ (see Appx. \S \ref{subsubsec:model_init}). 
Fig. \ref{fig:ablation_random}  shows that its AP-rank (median=87.5) is notably lower than that of random initialization (median=133). The difference, by one-sided Wilcoxon signed rank test, is statistically significant at $p=0.0204$.

\subsubsection{The Effect of Model Inclusion Criteria}
\begin{figure}[H]
\vspace{-0.05in}
\centering
	\includegraphics[width=0.85\linewidth]{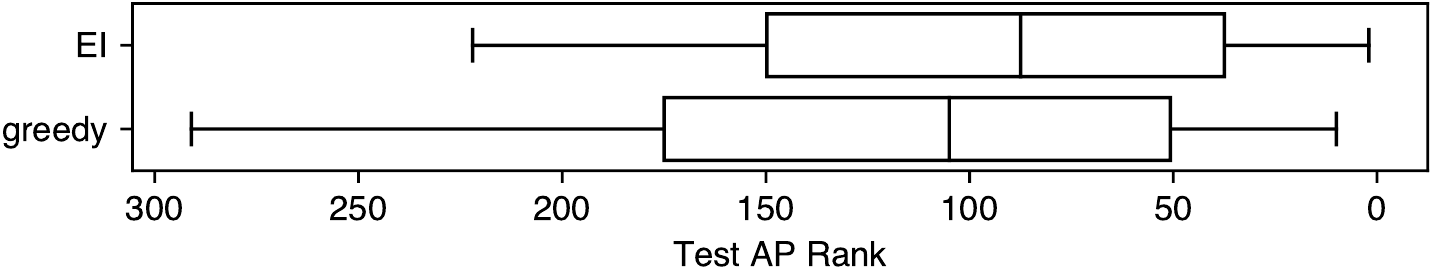}
	\vspace{-0.05in}
	\caption{Ablation of using EI (med.=87.5) vs. greedy without exploration (med.=105) during the adaptive search.
	}
	\label{fig:ablation_ei}
	\vspace{-0.05in}
\end{figure}

Fig. \ref{fig:ablation_ei} shows the performance comparison between using EI (balancing both exploitation and exploration) in Eq. (\ref{eq:EI}) and the greedy objective  
(exploitation only) that adds the model with the highest performance on $\mN$ during adaptive search (see \S \ref{subsubsec:adptive_search}). One-sided Wilcoxon signed rank test shows that the former (median=87.5) is statistically better (at $p=0.0203$) than the latter (median=105), justifying the use of EI.

\section{Related Work}
\label{sec:related_work}

\textbf{Model Selection for Outlier Detection.}
\label{subsec:model_selection_od}
There is no shortage of unsupervised 
OD algorithms in the literature, while 
most are 
sensitive to their hyperparameter (HP) choices \cite{Campos2016}.
Stunningly, 
outlier model selection (OMS) is 
understudied \cite{Bahri2022automl}. 

We can categorize the (short list of) approaches to OMS into two. The first group focuses on 
using \textit{internal} evaluation/performance measures based solely on model output 
and/or input data \cite{goix2016evaluate,marques2020internal}. As a recent study showed \cite{DBLP:journals/corr/abs-2104-01422}, those are noisy performance indicators that
are only slightly better than random selection.
The second category consists of learning-based approaches. A subset of these   
are semi-supervised \cite{li2020autood,li2020pyodds}, which use clean/inlier-only data for training, and
a small validation set \textit{with labels} for model selection---inapplicable 
to the U(nsupervised)OMS problem.
The only existing work on UOMS 
is the meta-learning based \metaod
\cite{DBLP:journals/corr/abs-2009-10606}. Distinctly, we eliminate the hand-crafted meta-features and instead employ performance-driven similarity between tasks. 
Consequently,
our  \method outperforms \metaod \cite{DBLP:journals/corr/abs-2009-10606} on  two separate testbeds.

\noindent
\textbf{
Hyperparameter Opt. (HPO) and  Meta-Learning.~}
\label{subsec:model_select_hyper_tuning}
HPO has gained significant attention within AutoML  owing to the advent of complex models with large HP spaces that are costly to train and thereby to tune \cite{feurer2019hyperparameter}. 
Besides 
model-free techniques such as grid or 
random search \cite{li2017hyperband},
Bayesian optimization and 
the adaptation of Sequential Model-Based Optimization (SMBO) \cite{DBLP:journals/jgo/JonesSW98} is one of the main lines of work in HPO. 
The idea is to iterate between (i) fitting a \textit{surrogate} performance function onto past HP evaluations, 
and (ii) using it to choose the next HP 
based on an \textit{acquisition} function. 
Well-established SMBO approaches  include SMAC \cite{hutter2011sequential} and Auto-WEKA \cite{thornton2013auto}. 
We remark that SMBO cannot directly be  used for UOMS, as 
we cannot evaluate model performance reliably to train an effective surrogate.
Instead, \method employs meta-learning to 
estimate the mean and variance of a candidate model's performance (to be used for \textit{acquisition}) based on similar historical tasks. 

Meta-learning has been used for HPO 
in various forms \cite{wistuba2016hyperparameter}; e.g.
to warm-start SMBO \cite{feurer2015initializing},  prune the HP search space \cite{wistuba2015hyperparameter}, and transfer surrogate models \cite{yogatama2014efficient}. 
Active testing \cite{conf/mldm/LeiteBV12} has used
meta-learning for model selection, 
which differs from our work in two key aspects. First, their acquisition 
is fully exploitative and does not factor in variance. 
More importantly, 
it is supervised: computes task similarity based on model performance evaluated on ground-truth. 

\section{Conclusion}
\label{sec:conclusion}

In the face of numerous outlier detection algorithms with various hyperparameters, there exists a  shortage of principled approaches to \textit{unsupervised} outlier model selection---a vastly  understudied subject. 
Toward filling this gap in the literature, 
we proposed \method, a  meta-learning approach that selects a candidate model for a new task based on its \textit{performance-based similarity} to historical (meta-train) tasks. \method adaptively identifies these neighbor tasks, as such, it can flexibly output a selected model in an any-time fashion, accommodating varying time budgets. Through extensive experiments, we showed that \method significantly outperforms a wide range of prior as well as more recent baselines. 
Future work 
includes extending to UOMS  for 
deep learning based outlier models.

\section*{Acknowledgments}{

This work is sponsored by NSF CAREER 1452425. We also thank PwC Risk and Regulatory Services Innovation Center at Carnegie Mellon University. Any conclusions expressed in this material are those of the author and do not necessarily reflect the views, expressed or implied, of the funding parties.

}

\vspace{-0.05in}
\bibliographystyle{IEEEtran}
\bibliography{ref}

\vspace{-0.1in}
\appendix
\setcounter{table}{0}

\section{Implementation Details}
\label{sec:implement}

\subsection{(Offline) Meta-training Details for \S \ref{sec:method}}
\label{subsec:train_details}

\subsubsection{Building Internal Performance Measures (IPMs)}
\label{subsubsec:internal_measure}

As described in \S \ref{ssec:ipm} and \ref{ssec:performance_predictor}, IPMs are used as the input features of the performance predictor $f(\cdot)$.
In \method, we use three consensus-based IPMs (i.e., MC, SELECT, and HITS) as they are reported to carry useful signals in model selection \cite{DBLP:journals/corr/abs-2104-01422}. 
Namely, consensus-based IPMs consider the resemblance to the overall consensus of outlier scores as a sign of a better model; their computation requires a group of models.

In \cite{DBLP:journals/corr/abs-2104-01422}, Ma \textit{et al.} use all models in $\mM$ for building IPMs, leading to high cost in generating outlier scores and then IPMs. To reduce the cost, we identify a small subset of representative models $\mM_A \in \mM$ called the \textit{anchor} set (i.e., $|\mM_A|\ll |\mM|$), for calculating IPMs. That is, we generate the IPMs of a model with regard to its consensus to $\mM_A$ rather than $\mM$.
Similar to forward feature selection \cite{guyon2003introduction}, the anchor set can be identified in a forward fashion (i.e., iteratively expanding the set) and cross-validation on the meta-train database.

\subsubsection{Pairwise Performance Predictor}
\label{subsubsec:regressor}
As shown in \S \ref{ssec:performance_predictor}, the predictor $f(\cdot)$ maps the vector of IPMs of a pair of models to their performance difference. To that end, we train a LightGBM regressor \cite{DBLP:conf/nips/KeMFWCMYL17} by enumerating all ${m \choose 2}$ model pairs for each task in meta-train database, where the hyperparameters are set by cross-validation.

\subsection{(Online) Model Selection Details for \S \ref{sec:online_model_selection}}
\label{subsec:prediction_details}

\subsubsection{Hyperparameters}
\label{subsubsec:hyperparameters}
The hyperparameters are chosen by LOOCV on the meta-train set. To that end, we find the following settings works well in both testbeds: (\textit{i}) the size of the neighbor set, $|\mN|=t$, equals to 5 (\S \ref{ssec:neighbors}) (\textit{ii}) the initial size of the model subset, $|\mM_s|$, equals to 7 (\S \ref{sssec:init}) and (\textit{iii}) patience $p$ equals to 17 (\S \ref{subsubsec:convergence}).

\subsubsection{Model Subset Initialization}
\label{subsubsec:model_init} Other than random sampling, 
we design a coverage-driven strategy for initializing $\mM_s$ in \S \ref{sssec:init}. Intuitively, we expect $\mM_s$ 
provides differentiability among datasets---the models' performance in $\mM_s$ on different datasets should vary. To this end, the coverage-driven strategy iteratively builds the initial $\mM_s$ by including the model that performs best (a top model) or worst (a bottom model) on the most meta-train tasks that have \textit{not} been ``covered". 
A task is said ``covered" if  both its top and bottom models (at least one) are already included in the model subset.

\subsection{Detailed Complexity Analysis}
\label{sec:complexity}

Suppose each task has $r$ samples and $d$ features on average, and the score computation of an OD model takes $C_{\text{train}}(r,d)$.

\noindent \textbf{Meta-training} involves
(\textit{i}) IPM generation (\S \ref{ssec:ipm}) for $n$ meta-train tasks on all $m$ models; MC, SELECT, and HITS have $O(nr)$, $O(nmr)$, $O(nmr)$, respectively;
(\textit{ii}) training of pairwise performance predictor (i.e., lightGBM, \S \ref{ssec:performance_predictor}) uses the input data composed by $n$ tasks, each with ${m \choose 2}$ model pairs by enumeration, leading to $O(nm^2)$ complexity; and (\textit{iii}) building  the anchor set with forward selection (\S \ref{subsubsec:internal_measure}) takes $O(nmr)$. Overall runtime is $O(nmr+nm^2)$.

\noindent \textbf{Model selection} first initializes the model subset $\mM_s$ with the coverage-driven strategy (\S \ref{sssec:init}), yielding $O(nm)$. For each initial model in $\mM_s$, outlier score and IPM generation take $O(C_{\text{train}}(r,d)+r)$. In each of $b$ iterations of adaptive search (\S \ref{subsubsec:adptive_search}), \method (\textit{i}) predicts the
model performance with $O(bn)$ complexity (\textit{ii}) identifies the next model to be included in $\mM_s$ with EI, taking $O(mt)$ and (\textit{iii}) gets the next model's outlier scores and IPMs with 
$O(C_{\text{train}}(r,d)+r)$ runtime. The total selection 
runtime is 
$O[b(bn+mt+C_{\text{train}(r,d)})]$. 

\label{sec:appendix}

\end{document}